# Cornell SPF: Cornell Semantic Parsing Framework

Yoav Artzi
Department of Computer Science and Cornell Tech
Cornell University
New York, NY 10011
yoav@cs.cornell.edu

The Cornell Semantic Parsing Framework (SPF) is a learning and inference framework for mapping natural language to formal representation of its meaning. The framework implements Combinatory Categorial Grammar (CCG; Steedman, 2000, 2011; Steedman and Baldridge, 2003) to map sentences to lambda calculus logical forms (Church, 1932, 1940; Montague, 1970a,b, 1973). The framework includes algorithms for learning and inference. SPF is implemented in Java and the code is available online.[1] The code repository includes a README and basic documentation. For a detailed introduction to formalism and implemented in SPF, see our tutorial.[2] If you use SPF, please cite:

```
@misc{Artzi:16spf,
    Author = {Yoav Artzi},
    Title = {Cornell {SPF}:  Cornell
Semantic Parsing Framework},
    Year = {2016},
    Eprint = {arXiv:1311.3011},
}
```

This document has been updated for version 2.0, and will be updated as new versions are released.

## 1 Projects Using SPF

This is a partial list of projects that have used SPF:[3]

- Bootstrapping Semantic Parsers from Conversations (Artzi and Zettlemoyer, 2011)

- Learning to Parse Natural Language Commands to a Robot Control System (Matuszek et al., 2012b)

- A Joint Model of Language and Perception for Grounded Attribute Learning (Matuszek et al., 2012a)

- Weakly Supervised Learning of Semantic Parsers for Mapping Instructions to Actions (Artzi and Zettlemoyer, 2013)

- Scalable Semantic Parsing with Partial Ontologies (Kwiatkowski et al., 2013)

- Learning Distributions over Logical Forms for Referring Expression Generation (FitzGerald et al., 2013)

- Learning Compact Lexicons for CCG Semantic Parsing (Artzi et al., 2014)

- Context-dependent Semantic Parsing for Time Expressions (Lee et al., 2014)

- Scalable Semantic Parsing with Partial Ontologies (Choi et al., 2015)

- Learning to Interpret Natural Language Commands through Human-Robot Dialog (Thomason et al., 2015)

- Broad-coverage CCG Semantic Parsing with AMR (Artzi et al., 2015)

- Neural Shift-Reduce CCG Semantic Parsing (Misra and Artzi, 2016)

---

[1] http://yoavartzi.com/spf
[2] http://yoavartzi.com/tutorial
[3] If you are using SPF for your work, please let us know, so we can update this list.

## 2 License

## 3 Acknowledgments

We are thankful for the support, advice, and guidance of Luke Zettlemoyer, Mark Steedman, Tom Kwiatkowski, Eunsol Choi, Dipanjan Das, Nicholas FitzGerald, Kenton Lee, Cynthia Matuszek, Dipendra Misra, Slav Petrov, Mark Yatskar, the University of Washington NLP Group, and the Cornell NLP Group. Work on SPF has been supported by a Microsoft PhD Fellowship, Google Faculty Award, and AWS Cloud Credits for Research.